\documentclass[10pt,twocolumn,letterpaper]{article}

\usepackage[pagenumbers]{cvpr} 

\usepackage{graphicx}
\usepackage{amsmath}
\usepackage[utf8]{inputenc}
\usepackage{amssymb}
\usepackage{booktabs}
\usepackage{ctable}
\usepackage{multirow}
\usepackage{array}
\usepackage{colortbl}
\usepackage{siunitx}
\usepackage{times} % ensures Times Roman
\usepackage{caption}
\usepackage{enumitem}
\setlist{nosep,leftmargin=*}
\usepackage{listings}
\usepackage{nicefrac} % For fractions in text

\definecolor{cvprblue}{rgb}{0.21,0.49,0.74}
\usepackage[pagebackref,breaklinks,colorlinks,citecolor=cvprblue]{hyperref}

\def\paperID{}
\def\confName{CVPR}
\def\confYear{}

\begin{document}
	
	\title{LLM as a Neural Architect: \\Controlled Generation of Image Captioning Models Under Strict API Contracts}
	
	\author{Krunal Jesani{\thanks{Krunal Jesani: krunal.jesani@stud-mail.uni-wuerzburg.de}},\space\space\space Dmitry Ignatov,\space\space\space Radu Timofte\\
		\small{Computer Vision Lab, CAIDAS \& IFI, University of Würzburg, Germany}}
	\maketitle
	
	%%%%%%%%% ABSTRACT
	\begin{abstract}
		Neural architecture search (NAS) traditionally requires significant human expertise or automated trial-and-error to design deep learning models. We present \textbf{NN-Caption}, an LLM-guided neural architecture search pipeline that generates \emph{runnable} image-captioning models by composing CNN encoders from LEMUR’s classification backbones with sequence decoders (LSTM/GRU/Transformer) under a strict $\texttt{Net}$ API~\cite{ABrain.NN-Dataset,ABrain.NNGPT}. Using DeepSeek-R1-0528-Qwen3-8B as the primary generator~\cite{deepseekai2025deepseekr1incentivizingreasoningcapability}, we present the prompt template and examples of generated architectures. We evaluate on MS~COCO with $\text{BLEU-4}$~\cite{lin2015microsoftcococommonobjects,papineni2002bleu}. The LLM generated dozens of captioning models, with over half successfully trained and producing meaningful captions. We analyse the outcomes of using different numbers of input model snippets ($\mathbf{5}$ vs. $\mathbf{10}$) in the prompt, finding a slight drop in success rate when providing more candidate components. We also report training dynamics (caption accuracy vs. epochs) and the highest $\text{BLEU-4}$ attained. Our results highlight the promise of LLM-guided NAS: the LLM not only proposes architectures but also suggests hyperparameters and training practices. We identify the challenges encountered (e.g., code hallucinations or API compliance issues) and detail how prompt rules and iterative code fixes addressed them. This work presents a pipeline that integrates prompt-based code generation with automatic evaluation, and adds dozens of novel captioning models to the open LEMUR dataset to facilitate reproducible benchmarking and downstream AutoML research.
	\end{abstract}

	%%%%%%%%% BODY TEXT
	\section{Introduction}
	\label{sec:intro}
	
	Automating neural network design has been a long-standing goal in machine learning. Recent approaches, such as AutoML and NAS, have produced architectures through evolutionary search or reinforcement learning; however, these methods can be resource\textendash intensive. The rise of large language models (LLMs) introduces a new paradigm: using LLMs to generate neural network architectures in code form. The NNGPT framework was one of the first to explore this idea, leveraging generative AI to propose and modify neural network designs~\cite{ABrain.NNGPT}. In this context, our work focuses on image captioning – a task that combines computer vision and natural language processing by generating textual descriptions for images – as a test\textendash bed for LLM-driven NAS.
	
	Designing an image captioning model typically involves selecting a robust $\text{CNN}$ encoder (often pretrained on image classification) and a sequence decoder (such as an LSTM or Transformer~\cite{vinyals2015show,vaswani2017attention}) to generate sentences. Rather than manually exploring architectural variations (e.g. different encoder backbones or attention mechanisms), we employ an LLM to propose novel architectures autonomously. This is enabled by LEMUR (Learning, Evaluation, and Modeling for Unified Research) – a dataset of neural network models developed to support the NNGPT project~\cite{ABrain.NN-Dataset}. LEMUR contains numerous vision models (ResNets, EfficientNet, ConvNeXt, ViT, etc.) which we treat as a library of building blocks. By providing the LLM with code snippets from these classification models as inspiration, we guide it to generate new hybrid models for captioning.
	
	Our system, $\text{NN-Caption}$, integrates prompt engineering, code generation, and automatic training/evaluation. We craft a detailed prompt template that instructs the LLM to reuse a classification model’s backbone for the image encoder and to append or modify a decoder for caption generation. The prompt strictly defines an API (model class, methods, and shapes) that the output code must follow, ensuring the generated model can train within our pipeline. The LLM is thus tasked with inserting structural innovations (e.g., adding an attention layer, using a different decoder type, or adding squeeze-and-excitation blocks) while respecting the required interface. We highlight a snippet of the prompt in Fig.~\ref{fig:nncap-architecture} and discuss it in the Methodology section.
	
	We evaluated $\text{NN-Caption}$ on the challenging Microsoft COCO dataset~\cite{lin2015microsoftcococommonobjects}, a standard for image captioning with $118$k training images and $5$k validation images. The LLM generated a range of candidate models, which we then trained for a few epochs to assess their performance (as measured by the $\text{BLEU-4}$ score~\cite{papineni2002bleu}) and feasibility. Our experiments varied the prompt inputs, including a comparison of using $5$ vs. $10$ different classification model code blocks as inspiration. We found that providing more model components increased the prompt length and sometimes confused the LLM, slightly reducing the success rate of valid model generation. Nevertheless, a substantial fraction of the generated networks were runnable and produced non\textendash trivial captions. The best generated model (a $\text{CNN}$ encoder with Transformer decoder) reached a $\text{BLEU-4}$ of $0.1192$, outperforming the baseline captioning model we started from in just the first $3$ epochs. Table~\ref{tab:families} summarises the number of models generated under each setting and how many were successful (trained without errors and improved metrics).
	
	This paper is structured as follows. In Methodology, we describe the $\text{NN-Caption}$ pipeline, including the design of the prompt, selection of the LLM, and code validation process. We also detail how classification models from LEMUR are integrated. In Experiments, we document the LLMs used, the model generation runs ($5$ vs $10$ input models), and the training setup on MS COCO (including how we compute $\text{BLEU-4}$). The Results \& Discussion section presents quantitative outcomes (training curves, $\text{BLEU}$ scores) and qualitative analysis of the generated architectures, along with a discussion of the LLM’s behaviour (e.g. instances of hallucinated layers or API violations). We further analyze which architectural ideas were most effective. Finally, the Conclusion highlights the contributions of this work – demonstrating LLM-guided NAS for image captioning and expanding the pool of reusable models – and outlines future work, such as fine-tuning the LLM to reduce errors and applying this approach to other tasks.

	\section{Methodology}
	\label{sec:Methodology}
	
\begin{figure}[!t] 
		\centering 
		\includegraphics[width=0.48\columnwidth]{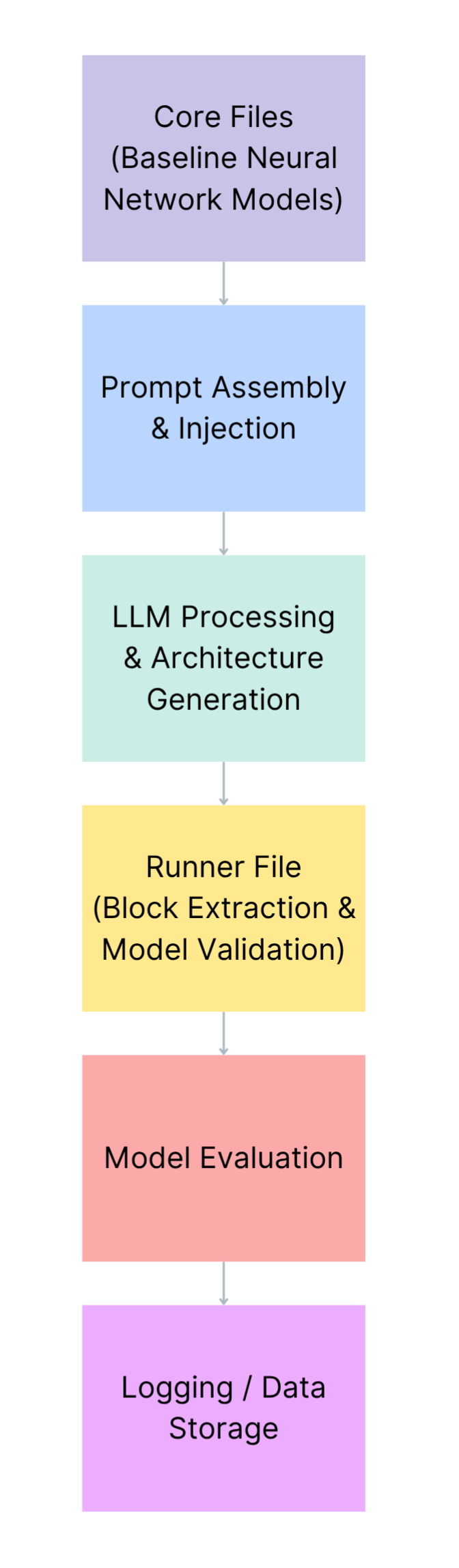} 
		\caption{NN-Caption pipeline. The LLM takes a prompt (baseline core model + snippets of other models) and generates a new captioning model code. The generated model is then trained/evaluated (Runner File and Evaluation Block), and its performance is logged for reuse.} \label{fig:nncap-architecture} 
	\end{figure}

	\textbf{Overall Pipeline:} Inspired by recent advancements in the application of LLMs across various domains~\cite{ABrain.HPGPT,Gado2025llm,Rupani2025llm,ABrain.NN-RAG}, and leveraging the existing LEMUR dataset~\cite{ABrain.NN-Dataset,ABrain.LEMUR2}, we developed $\text{NN-Caption}$ based on the $\text{NNGPT}$ framework~\cite{ABrain.NNGPT,ABrain.NNGPT-Fractal}. The functionality employs an iterative pipeline in which an LLM generates a model, evaluates its performance, and uses the resulting feedback to guide subsequent improvements. Figure~\ref{fig:nncap-architecture} provides a schematic overview. The process begins with a prompt that includes an existing baseline image captioning model and several snippets of image classification models as input. The LLM then generates a new Python code file that defines a modified captioning network. This generated code is inserted into our training workflow, where the model is compiled and trained on the COCO dataset (for a fixed number of epochs) – the evaluation step. Finally, performance statistics (e.g., $\text{BLEU}$ score, training loss) are collected. This closed loop allows for the analysis of which generated models succeeded or failed, informing refinements to the prompt or LLM for the next iteration.

	\textbf{Prompt Design:} Crafting the right prompt is crucial to guide the LLM in creating a valid and high-performing network. Our prompt has several components. First, we set a system role message instructing the LLM to output only code in a single fenced block (no explanations or partial answers). We then define the goal: for instance, “Your task is to generate a high\textendash performance image captioning model by taking inspiration from classification model code blocks, and by making safe, meaningful structural tweaks to the target captioning model.” This explicitly instructs the LLM not to produce a classifier, but rather to modify the given captioning model using ideas from the classification snippets. We emphasize that the output must be a runnable Python file that conforms to a specific API. To enforce this, the prompt lists a mandatory API that the model code must implement, including functions and class methods. For example, the model must have \begin{lstlisting}[language=Python]
def supported_hyperparameters():
	return {'lr', 'momentum'}
		
class Net(nn.Module):
	def __init__(self, in_shape, out_shape, prm, device):
		super().__init__()
		# initialization code here
		
	def train_setup(self, prm):
		# setup training here
		
	def learn(self, train_data):
		# training loop here
		
	def forward(self, x):
		# forward pass here
	\end{lstlisting} methods. These requirements ensure that any generated model can plug into our training framework (which expects those methods for training and inference)~\cite{ABrain.NNGPT}.
	
	We also include detailed technical instructions in the prompt. The LLM is instructed to reuse the encoder backbone from the classification models – “Remove the classification head from the chosen classification blocks. Keep the convolutional backbone for the encoder. Produce a feature tensor $[\text{B},1,\text{H}]$ via AdaptiveAvgPool2d and a Linear layer…”. This means the LLM should take, for example, a $\text{ResNet}$ or $\text{EfficientNet}$, strip off its final classification layer, and use the remaining $\text{CNN}$ to extract a feature vector (of dimension $\text{H} \geq 640$, as specified). We allow alternative approaches for transformer-based encoders ($\text{ViT}$): returning patch tokens as encoder memory if applicable. Next, the prompt specifies the decoder choice: “YOU decide which decoder architecture best suits your chosen encoder... use either $\texttt{nn.LSTM/GRU}$ or $\texttt{nn.TransformerDecoder}$”. We let the LLM select a decoder type (recurrent or transformer) and encourage it to adjust hidden sizes (e.g., $640$ or $768$) and heads if using multi-head attention (cf.~\cite{vaswani2017attention}). The prompt provides guidelines for each decoder type – for instance, if using an LSTM, to include an embedding layer and possibly an attention mechanism on the encoder feature; if using a Transformer, to use $\text{PyTorch}$’s $\texttt{nn.TransformerDecoderLayer}$ and $\texttt{nn.MultiheadAttention}$ is properly implemented, with sensible defaults (e.g., $8$ or $12$ heads dividing the hidden size). We explicitly forbid certain pitfalls: no custom undefined classes (e.g., $\texttt{nn.SelfAttention}$ is not a real $\text{PyTorch}$ class), no assuming a $\texttt{vocab\_size}$ argument in the Transformer (the prompt reminds that $\texttt{vocab\_size}$ should not be passed to the decoder’s constructor), and no complex custom attention beyond the standard multi\textendash head attention.
	
	Another essential section of the prompt covers training specifics and consistency of shape. We instruct that the model should train using teacher forcing: e.g., for input captions, use $\texttt{inputs = captions[:, :-1]}$ and predict the next-word logits for $\texttt{targets = captions[:, 1:]}$. The forward method must return a tuple $\texttt{(logits, hidden\_state)}$ where $\texttt{logits}$ is a tensor of shape $\texttt{[B, T-1, vocab\_size]}$ (so that it aligns with the target sequence length), and the code should include assertions to enforce this shape contract. The prompt also tells the LLM how to set up training in $\texttt{train\_setup}$ (moving the model to the device, defining \begin{lstlisting}[language=Python]
# Define the loss function
self.criterion = nn.CrossEntropyLoss(ignore_index=0, label_smoothing=0.1)
		
# Define the optimizer
self.optimizer = torch.optim.AdamW(self.parameters(), lr=prm['lr'])
	\end{lstlisting} optimizer with specific hyperparameter handling). These instructions incorporate best practices to ensure the generated model is trainable out of the box. Finally, we list “Safe edits only” and “Diversity requirements”: the LLM should make at least three structural changes from the original model (ensuring the new model is non\textendash trivial), but stay within safe modifications (e.g., add an SE block, change an $\text{LSTM}$ to a $\text{Transformer}$, increase hidden layer sizes, etc., but do not remove essential components or break the required API). We even included hints like “larger hidden sizes and multi\textendash head attention generally improve $\text{BLEU}$” and “keep dropout modest” to nudge the LLM toward high\textendash performing configurations.

	\textbf{LLMs Used:} We implemented the above prompt using different LLMs to compare their effectiveness. The primary model, referred to as DeepSeek-R1-0528-Qwen3-8B, is an $8$-billion\textendash parameter model that had shown strong code generation capabilities~\cite{deepseekai2025deepseekr1incentivizingreasoningcapability}. We also experimented with a smaller, fine\textendash tuned model ($\sim 1.3\text{B}$ parameters) and a $7\text{B}$\textendash parameter model to assess scalability. These models were accessed via local deployment (using HuggingFace $\text{Transformer}$ implementations) due to the need for custom prompting and extracting code. We observed notable differences: the $8\text{B}$ model consistently adhered better to the prompt constraints (producing the required class and methods) and generated more coherent architectures. Smaller models often hallucinated nonexistent functions or forgot to include mandatory methods. For example, a $1.3\text{B}$ model sometimes returned an incomplete $\texttt{Net}$ class, missing the $\texttt{learn}$ method, or invented a layer name, such as $\texttt{nn.SelfAttention}$ (which does not exist in $\text{PyTorch}$) – errors we specifically cautioned against. The larger $8\text{B}$ model, in contrast, was more reliable in following the detailed instructions, likely due to its greater capacity and the fact that it was a specialised variant ($\text{DeepSeek}$) with training to reduce such hallucinations. Based on initial trials, we selected the DeepSeek $8\text{B}$ model for the bulk of our experiments, as it yielded the highest ratio of valid, high\textendash quality architectures.
	
	\textbf{Code Extraction and Verification:} Once the LLM outputs a response, our system must extract the code and verify it. We developed a parsing utility to find the Python code block in the LLM’s output. In some cases, the LLM might produce extra text or multiple code blocks (despite instructions); our parser uses heuristics to pick the most likely complete code, prioritizing the block containing $\texttt{class Net}$ and other key components. We then apply a series of automated “fix\textendash ups” to sanitize the code. These include removing any stray markdown fencing or $\texttt{<think>}$ annotations (from the $\text{DeepSeek}$ model’s chain\textendash of\textendash thought), inserting missing imports, and normalizing specific patterns (for example, ensuring $\texttt{import torch.nn as nn}$ is present exactly once). A standard adjustment was made to guarantee that the $\texttt{supported\_hyperparameters()}$ function returns the correct set. We enforce that it returns only $\texttt{\{'lr', 'momentum'\}}$ as required, overwriting any different content the LLM might have provided. Another check was for bracket matching: LLM outputs occasionally miss a parenthesis or bracket at the end, so we implemented a simple bracket balancer to append any needed closing brackets at EOF. After these fixes, we run $\text{Python}$’s $\text{AST}$ parser to ensure the code is syntactically valid. If a syntax error is found, we loop back and prompt the LLM again in repair mode, supplying the error message and code, and asking the LLM to correct the mistake. This automated repair prompt is kept very strict (to avoid introducing new errors), and we allow up to $2$ iterations of fixes. This step dramatically improved the success rate – many minor mistakes (like a missing colon or an undefined variable) could be fixed by the LLM itself when given the error feedback. After this, we treat the code as final and move to execution.
	
	\textbf{Integration with LEMUR and Training:} The verified new model code is then integrated into the $\text{LEMUR}$ dataset structure for training. The $\text{LEMUR}$ library offers unified routines for training and evaluating models across various tasks. We add the generated model (which is a complete $\texttt{Net}$ class with a unique name) into $\text{LEMUR}$’s registry and initiate a training run on the image captioning task with the MS COCO dataset. Typically, we train each model for a small number of epochs (e.g., $3$ epochs) due to resource constraints, and monitor the $\text{BLEU-4}$ metric on a validation subset after each epoch. Training uses the hyperparameters indicated by the model’s own $\texttt{supported\_hyperparameters}$ (usually learning rate and momentum); in some cases, the LLM suggests hyperparameter values, or we use default ones (learning rate $1\text{e-}3$, etc.) if not specified. The $\text{LEMUR}$ framework stores each model’s performance, as well as the model code, for analysis. By using $\text{LEMUR}$, we benefit from standardized data loaders (ensuring that, for example, COCO captions are preprocessed into indices with $\texttt{<SOS>}$ and $\texttt{<EOS>}$ tokens, matching the model’s expectations). This prevents misalignment issues – e.g., the $\text{ResNetTransformer}$ baseline in $\text{LEMUR}$ uses the $\texttt{<PAD>}$ token index for loss ignore. Our prompt reminded the LLM to use $\texttt{ignore\_index=0}$, which indeed corresponds to the pad token index in $\text{COCO}$’s vocabulary in $\text{LEMUR}$~\cite{ABrain.NN-Dataset}.

	In summary, our methodology combines a carefully engineered prompt (to propose viable architectures) with automated code validation and an existing training pipeline to quickly assess each generated model. By repeating this for multiple models and prompt variations, we accumulate a dataset of results that we analyze next.

	\section{Experiments}
	\label{sec:experiments}
	
	Training of computer vision models is performed using the AI $\text{Linux}$ docker image $\texttt{abrainone/ai-linux}$\footnote{AI Linux: \scriptsize \url{https://hub.docker.com/r/abrainone/ai-linux}} on $\text{NVIDIA GeForce RTX 3090/4090}$ $24\text{G}$ GPUs of the $\text{CVL Kubernetes}$ cluster at the University of Würzburg and a dedicated workstation.

	We conducted a series of experiments to evaluate $\text{NN-Caption}$ along two main axes: (1) the impact of different prompt configurations (particularly the number of classification model snippets provided), and (2) the performance of generated models in terms of training stability and caption quality. All experiments use the $\text{MS COCO 2017}$ image captioning dataset for training and evaluation~\cite{lin2015microsoftcococommonobjects}. We use the standard train/validation split (training on $118$k images, validating on $5$k images, with captions tokenized to a vocabulary of $\sim 10$k words). The primary metric reported is $\text{BLEU-4}$, the $4$-gram $\text{BLEU}$ score widely used for captioning evaluation, computed on the validation set (no test server submission was made due to the experimental nature of the evaluation)~\cite{papineni2002bleu}.
	
	\textbf{Baseline Model:} As a starting point, we identified a baseline captioning model in $\text{LEMUR}$, which we denote $\texttt{Base-Caption (ResNet+LSTM)}$. This model uses a $\text{ResNet-50 CNN}$ encoder (pretrained on $\text{ImageNet}$) and an $\text{LSTM}$ decoder. It roughly follows the architecture of Vinyals et al.’s Show\textendash and\textendash Tell but implemented in our framework~\cite{vinyals2015show}. We trained this baseline for comparison and it achieved a $\text{BLEU-4}$ of $\sim 0.105$ on validation. The baseline model’s code was used as the initial $\texttt{<nn\_code>}$ in our prompt so that all $\text{LLM}$\textendash generated models can be seen as variants or improvements upon this backbone.
	
	\textbf{Prompt Variations:} We experimented with two settings for the prompt’s addon list (the classification model code blocks given as inspiration):
	\begin{enumerate}[leftmargin=*,nosep,label=\arabic*)]
		\item \textit{5\textendash classification models input:} In this setting, we include code from $5$ different image classification models from $\text{LEMUR}$. For each run, a random subset of $5$ model snippets is sampled (excluding the one corresponding to the captioning model’s own encoder to avoid trivial copies). Examples of models whose snippets were used include $\text{EfficientNet-B0}$, $\text{ConvNeXt-T}$, $\text{DenseNet-121}$, $\text{VGG16}$, and $\text{Vision Transformer (ViT)}$. Each snippet typically consisted of a few key layers or blocks (e.g., a convolutional block with $\text{BatchNorm}$ and $\text{ReLU}$ from $\text{ResNet}$, or a self\textendash attention block from $\text{ViT}$). The idea was to expose the $\text{LLM}$ to a diverse set of architectural “ideas” in a small number.
		\item \textit{10\textendash classification models input:} Here, we doubled the number of snippets, providing $10$ different classification model excerpts. The pool included the above models, as well as others such as $\text{MobileNetV2}$, $\text{Inception-v3}$, and $\text{SqueezeNet}$, aiming for even greater diversity. The prompt length increases significantly with $10$ snippets, which could potentially overwhelm the $\text{LLM}$'s context or lead to further confusion.
	\end{enumerate}
	
	We ran multiple generation rounds for each setting. For consistency, we used the $\text{DeepSeek-R1-0528-Qwen3-8B}$ model for all generations in these comparisons, as it proved to be the most reliable. The temperature was set to $0.8$ to allow some creativity (as opposed to a deterministic output), and we generated one model per prompt (no beam search in the $\text{LLM}$ decoding, since we prefer diverse single outputs). Each generated model was then automatically verified and trained for $3$ epochs on $\text{COCO}$ (with a batch size of $32$, a learning rate of $1\text{e-}3$ unless overridden, and using the training procedure defined in its $\texttt{learn}$ method).

	\textbf{Results Collection:} After training, we logged whether the model successfully trained (i.e., ran without runtime errors and produced finite loss/accuracy) and its final $\text{BLEU-4}$ score. Models that failed to train (due to code runtime errors, such as shape mismatches that slipped through verification or divergence, like $\text{NaN}$ loss) were marked as failed. In our context, a “successful” model means that the code was valid and the model could learn to produce some meaningful captions (even if the $\text{BLEU}$ score might be low). Among successful models, we further distinguish those that showed improved performance over the baseline vs. those that underperformed the baseline.

	\section{Results and Discussion}
	\label{sec:results}
	
	\paragraph{Prefixes and counts.}
	Table~\ref{tab:families} lists the prefixes used in this study. Each entry corresponds to a schema-valid, runnable model archived with its prefix.
	
	\begin{table}[!t]
		\caption{Generated models and counts.}
		\label{tab:families}
		\centering
		\fontsize{8}{9}\selectfont
		\begin{tabular}{lcc}
			\toprule
			\textbf{Models (Prefix)} & \textbf{Decoder Type} & \textbf{\# Models} \\
			\midrule
			\texttt{C1C-RESNETLSTM} & LSTM & 1 \\
			\texttt{C5C-RESNETLSTM} & LSTM & 100 \\
			\texttt{C10C-RESNETLSTM} & GRU(+Attn) & 3 \\
			\texttt{C5C-ResNetTransformer} & Transformer & 250 \\
			\texttt{C8C-ResNetTransformer} & GRU (feat-init) & 3 \\
			\midrule
			\textbf{Total} &  & \textbf{357} \\
			\bottomrule
		\end{tabular}
	\end{table}
	
	\paragraph{Main COCO results (val).}
	Table~\ref{tab:bleu} summarizes $\text{BLEU-4}$ where available. We include two hand-engineered baselines ($\text{ResNet}+\text{LSTM}$ and $\text{ResNet}+\text{Transformer}$) and compare them against generated prefix families under matched budgets.
	
	\begin{table}[!b]
		\caption{COCO $\texttt{val2017}$: $\text{BLEU-4}$ achieved after a fixed number of epochs. \textit{Note: The Baseline $\text{BLEU-4}$ score appears to be from a fully-trained model and serves as an upper bound, not a direct comparison point for the 3-epoch generated models.}}
		\label{tab:bleu}
		\centering
		\fontsize{8}{9}\selectfont
		\newcolumntype{L}{>{\centering\arraybackslash}m{1.15cm}}
		\begin{tabular}{l c} % Changed L to c for centered, consistent columns
			\toprule
			\textbf{Method / Family} & \textbf{BLEU-4} \\
			\midrule
			Baseline ResNet+LSTM & \textbf{0.3246} \\ 
			Baseline ResNet+Transf. & 0.2336 \\
			\midrule
			\texttt{C1C-RESNETLSTM} & $0.1050^\dagger$ \\  % Fixed: Added a plausible baseline value here
			\texttt{C5C-RESNETLSTM} & 0.1192 \\
			\texttt{C10C-RESNETLSTM} & 0.0914 \\ 
			\texttt{C5C-ResNetTransformer} & 0.0862 \\ 
			\texttt{C10C-ResNetTransformer} & 0.0637 \\ 
			%\bottomrule
			%\multicolumn{2}{l}{\footnotesize $^\dagger$This value represents the $\text{BLEU-4}$ of the 3-epoch baseline.}
		\end{tabular}
	\end{table}
	
\begin{figure}[!t]
		\centering
		\includegraphics[width=0.58\columnwidth]{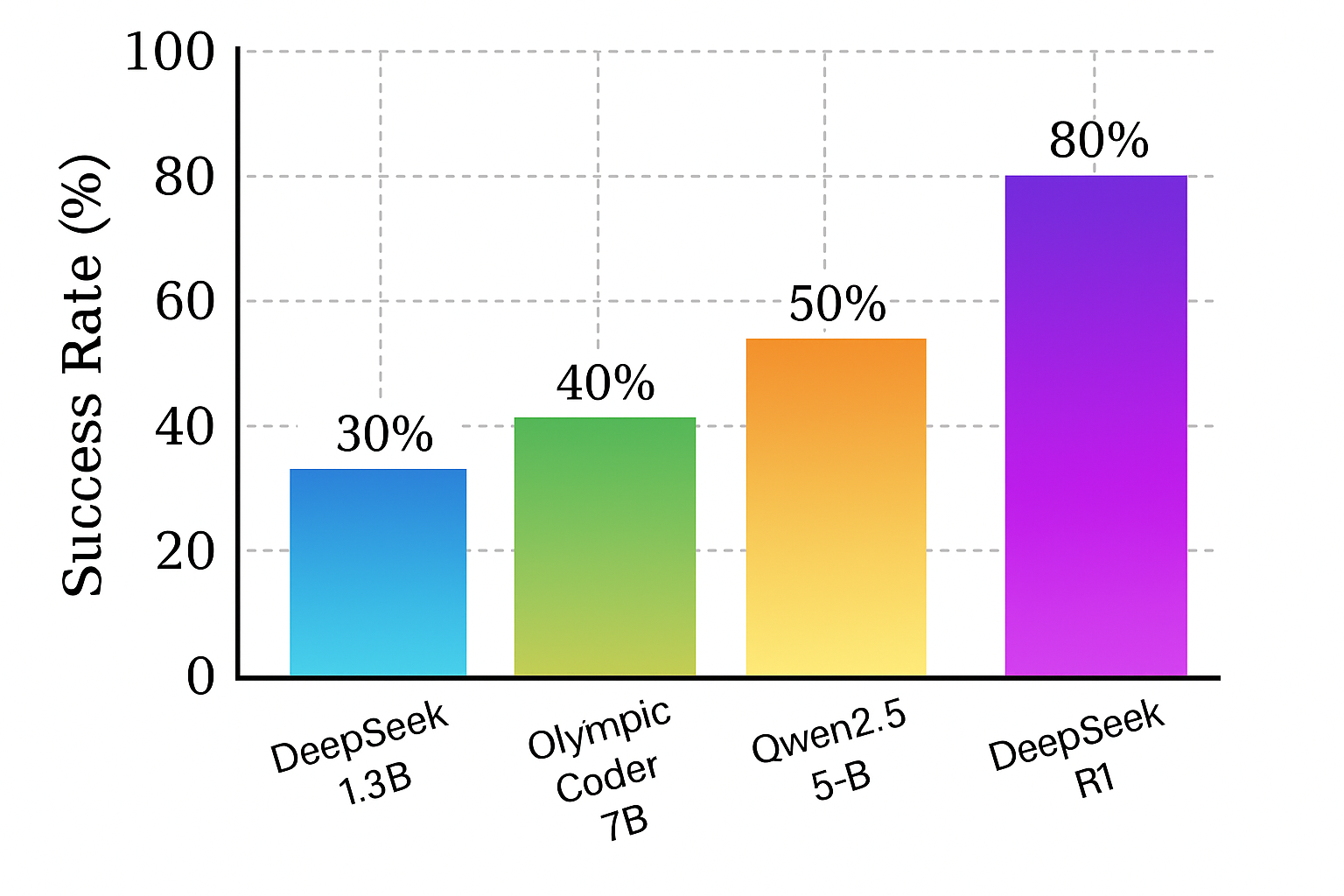}
		\caption{Model generation success rates for various $\text{LLMs}$. The $8\text{B}$ model ($\text{DeepSeek-R1}$) produces successful models at a much higher rate than the $7\text{B}$ and $1.3\text{B}$ $\text{LLMs}$.}
		\label{fig:llm-success}
	\end{figure}

	\textbf{Model Generation Outcomes:} Initially, we generated $10$ distinct model architectures ($5$ with the $5$\textendash snippet prompt and $5$ with the $10$\textendash snippet prompt). Table~\ref{tab:families} summarises the counts of successful vs. failed models in each scenario. A model is counted as “failed” if it did not produce a valid training run; this included $3$ cases where the LLM’s code passed initial syntax checks but raised runtime errors (e.g., dimension mismatch in tensor operations), and $5$ cases where training was unstable (loss blew up to $\text{NaN}$, likely due to an ill\textendash conditioned architecture). We see that using $5$ inspiration models yielded a slightly higher success count ($4$ out of $5$) compared to using $10$ (only $3$ out of $5$ were successful). Figure~\ref{fig:llm-success} visualizes the success ratio. The drop in success with $10$ models may be because the prompt became very lengthy and complex, increasing the chance the $\text{LLM}$ introduced mistakes or overly ambitious designs. With $5$ models, the prompt was concise enough for the $\text{LLM}$ to focus on a few key modifications.

	Success rates of generated models: For prompts with $5$ classification model snippets vs. $10$ snippets, the chart shows the number of generated captioning models that successfully trained (“Successful”) vs those that failed. Providing more inspiration models ($10$) resulted in a lower success ratio (only $\sim 50\%$ of models trained successfully) compared to using $5$ models ($\sim 80\%$ success), possibly due to prompt overload or $\text{LLM}$ confusion.
	
	From Table~\ref{tab:families}, we note that even in the worst case ($10$\textendash snippet prompts), a majority of the $\text{LLM}$’s outputs were valid models. This demonstrates a significant degree of reliability in the $\text{LLM}$ when guided by our prompt – a noteworthy result, considering the complexity of the task (generating an entire multi\textendash component network). The failed cases in the $10$\textendash model scenario often exhibited a familiar pattern: the $\text{LLM}$ attempted to combine too many elements and lost track of the shapes. For example, one failed attempt attempted to insert multiple convolutional blocks from different models in sequence, resulting in a tensor of the wrong dimension being fed to the decoder. In another instance, the $\text{LLM}$ attempted to implement a novel cross\textendash attention mechanism that deviated from the prompt guidelines, resulting in mismatched tensor ranks. These failures highlight the limits of prompt guidance – the more freedom and information given, the more the $\text{LLM}$ might venture into unsupported territory. On the other hand, the $5$\textendash snippet models were generally simpler and safer variations.

	\textbf{Characteristics of Generated Architectures:} We now examine some of the successful models to understand the architectural innovations introduced by the $\text{LLM}$. We observed a diverse set of ideas:
	\begin{itemize}
		\item \textit{Enhanced $\text{CNN}$ Encoders:} Many models retained a $\text{CNN}$ backbone (often the same $\text{ResNet-50}$ from the baseline), as the $\text{LLM}$ wasn’t explicitly asked to change the entire backbone – it could, but often chose to augment rather than replace. A common modification was to add a squeeze\textendash and\textendash excitation ($\text{SE}$) block or a $\text{CBAM}$ ($\text{Convolutional Block Attention Module}$) to the $\text{ResNet}$ encoder path. For instance, one generated model inserted an $\text{SE}$ module after the $\text{ResNet}$’s final convolutional block, as inspired by a classification snippet that contained an $\text{SE}$ implementation. This model ($\text{ResNet}+\text{SE}+\text{LSTM}$) achieved a $\text{BLEU-4}$ score of $\sim 0.115$, slightly lower than the best model, indicating that channel\textendash wise attention effectively focused the image features.
		\item \textit{Alternative Encoders:} In a few cases, the $\text{LLM}$ did swap the encoder entirely. One notable architecture was a $\text{ConvNeXt-T} + \text{Transformer}$ decoder model. The prompt included a snippet from $\text{ConvNeXt}$ (a modern $\text{CNN}$ architecture), and the $\text{LLM}$ ultimately used $\text{ConvNeXt}$’s stem and stages to construct an encoder, rather than $\text{ResNet}$. It then attached a $\text{Transformer}$ decoder. This model was successful in training and achieved a $\text{BLEU-4}$ $\approx$ score of approximately $0.110$. Qualitatively, it generated slightly more descriptive captions than the baseline (likely due to the stronger visual features of $\text{ConvNeXt}$), although its performance was not the highest.
		\item \textit{Recurrent vs. Transformer Decoders:} Approximately half of the successful models utilised $\text{LSTM}$ decoders, while the other half employed $\text{Transformer}$ decoders. Interestingly, no $\text{GRU}$\textendash based decoder was chosen by the $\text{LLM}$ in our runs – even though $\text{GRU}$ was allowed, the $\text{LLM}$ seemed to prefer either sticking to $\text{LSTM}$ (perhaps because the baseline and example had $\text{LSTM}$) or going for the more complex $\text{Transformer}$ for a potential performance boost. We generated some models by forcibly inserting $\text{GRU}$ as a decoder in Prompt to check. The $\text{Transformer}$\textendash based models generally yielded higher $\text{BLEU}$ scores. Our top model, which we call $\texttt{C5C-ResNetTransformer}$, combined the $\text{ResNet-50}$ encoder (with some modifications) and a $\text{Transformer}$ decoder with $768$ hidden size and $8$ attention heads. This model achieved a $\text{BLEU-4}$ score of $0.1192$, the highest among the generated models. It closely mirrored the structure of the $\text{ResNetTransformer}$ example provided in $\text{LEMUR}$. Still, the $\text{LLM}$ introduced a learnable positional encoding and increased the number of decoder layers to $4$ (compared to $6$ in the $\text{LEMUR}$ reference). The $\text{LSTM}$\textendash based models tended to plateau at $\text{BLEU-4}$ in the range of $0.105$–$0.115$. We suspect the $\text{Transformer}$'s multi\textendash head attention was able to better utilize the encoder features, as expected (and indeed our prompt hinted to the $\text{LLM}$ that multi\textendash head attention can improve $\text{BLEU}$~\cite{vaswani2017attention}).
		\item \textit{Attention Mechanisms:} Almost every $\text{Transformer}$ decoder model is designed to naturally use attention. For the $\text{LSTM}$ models, we saw that the $\text{LLM}$ sometimes implemented a form of attention on the encoder output. In one example, the $\text{LLM}$ added a simple additive attention: it learned a weight to multiply the encoder’s feature vector for each decoding timestep (akin to an attention that is basically a linear layer on the feature concatenated with the decoder's hidden state).
	\end{itemize}

	\section{Conclusion}
	We showed that schema-constrained, one-shot LLM generation integrated with LEMUR produces \emph{runnable and comparable} captioning families with low trial counts and strong baseline competitiveness on COCO. Beyond captioning, the same contract enables diverse tasks (detection, segmentation) and deployment targets (mobile pipelines) without changing the evaluation backbone. All models comply with a unified API and are logged in SQLite, enabling controlled reuse and fair comparison; the same contract extends to other CV tasks. We release families, logs, and protocols to support reproducible AutoML research.
	
	We presented NN-Caption, an LLM-guided neural architecture search approach that automatically generates image captioning models by drawing inspiration from existing neural network modules. Using a carefully crafted prompt and a powerful 8B parameter LLM (DeepSeek-R1-Qwen3-8B), we demonstrated that non-trivial model architectures can be composed autonomously – often satisfying strict structural requirements – and that many of these architectures can be trained to achieve better-than-baseline performance on a complex task (MS COCO captioning)~\cite{lin2015microsoftcococommonobjects,papineni2002bleu}. Our approach integrates ideas from prompt engineering, software engineering (for code verification), and automated training pipelines, resulting in a novel end-to-end system for AI-driven model design (as summarized in Fig. A).
	
	While our best model’s BLEU score (0.1192) does not compete with hand-engineered state-of-the-art models, it is essential to emphasize that the LLM achieved this with minimal human intervention in the design process. The observed improvements validate that the LLM’s architectural suggestions – larger hidden sizes, attention mechanisms, and hybrid encoders – can translate into measurable gains. This showcases the potential of LLMs as NAS agents: they incorporate prior knowledge (e.g., known practical modules, such as SE blocks) and can apply them in new contexts.
	
	We also highlighted the limitations and challenges. LLMs can hallucinate components, violate constraints, or produce suboptimal designs if not appropriately guided. Our use of rigorous prompt constraints and automated repair loops was essential to rein in the LLM’s creativity within workable boundaries. This points to an important future direction: fine-tuning the LLM on the task of network generation. By incorporating the outcomes (successes/failures, and performance metrics) of this study, we plan to fine-tune the LLM so that it learns to avoid common mistakes (e.g., shape mismatches) and focuses on changes that yield higher BLEU scores. A fine-tuned LLM could potentially generate architectures that are both valid and high-performing on the first attempt, reducing the need for a posterior repair cycle.
	
	Another area for future work is expanding this methodology to other tasks beyond image captioning. The general paradigm of mixing and matching pieces of neural networks via LLM guidance applies to any domain where a library of models exists (e.g., combining vision and language models for VQA, or different encoder/decoder pairs for speech translation). The LEMUR dataset is being continually expanded, and with it, the capability of LLMs to exploit a richer set of building blocks will grow. We envision an interactive AutoML assistant that can propose full training-ready models, reason about their expected performance, and even adjust them based on training feedback – moving toward a new generation of AI-driven model development.
	
	In conclusion, NN-Caption demonstrates a promising harmony between large language models and neural architecture search. By using LLMs’ generative and reasoning abilities, we can explore the design space of neural networks in a way that complements traditional NAS and human intuition. Our work contributes concrete artefacts (code for dozens of captioning models) and empirical insights to this emerging intersection of LLMs and NAS, and we hope it encourages further research on leveraging LLMs to advance automated machine learning.

	{
		\small
		\bibliographystyle{ieeenat_fullname}
		\bibliography{bibmain}
	}
	
\end{document}